\documentclass{article} % For LaTeX2e
\usepackage{iclr2017_workshop,times}
\usepackage{hyperref}
\usepackage{url}
\usepackage{amsmath}
\usepackage{amssymb}
\usepackage{graphicx}
\usepackage{mathtools}
\usepackage{xcolor}
\usepackage{algorithm}
\usepackage{algpseudocode}
\usepackage{graphicx}
\usepackage{caption}
\usepackage{subcaption}
\usepackage{wrapfig}
\usepackage{varwidth}
\usepackage{siunitx}

\newcommand{\R}{\mathbb{R}}
\newcommand{\numagents}{K}
\DeclareMathOperator{\sign}{sgn}
\DeclareMathOperator*{\expect}{\mathbb{E}}
\DeclareMathOperator*{\proba}{\mathbb{P}}
\DeclareMathOperator*{\var}{var}

\title{Particle Value Functions}

\author{Chris J. Maddison\textsuperscript{1,2}, Dieterich Lawson\textsuperscript{3}, George Tucker\textsuperscript{3},\\ 
{\bf Nicolas Heess\textsuperscript{2}, Arnaud Doucet\textsuperscript{1}, Andriy Mnih\textsuperscript{2}, Yee Whye Teh\textsuperscript{1,2}}\\
\textsuperscript{1}University of Oxford, \textsuperscript{2}DeepMind, \textsuperscript{3}Google Brain\\
\texttt{cmaddis@stats.ox.ac.uk} \\
}

\begin{document}
\maketitle

\begin{abstract}
The policy gradients of the expected return objective can react slowly to rare rewards. Yet, in some cases agents may wish to emphasize the low or high returns regardless of their probability. Borrowing from the economics and control literature, we review the risk-sensitive value function that arises from an exponential utility and illustrate its effects on an example. This risk-sensitive value function is not always applicable to reinforcement learning problems, so we introduce the \emph{particle value function} defined by a particle filter over the distributions of an agent's experience, which bounds the risk-sensitive one. We illustrate the benefit of the policy gradients of this objective in Cliffworld.
\end{abstract}

\section{Introduction}

The expected return objective dominates the field of reinforcement learning, but makes it difficult to express a tolerance for unlikely rewards. This kind of risk sensitivity is desirable, e.g., in real-world settings such as financial trading or safety-critical applications where the risk required to achieve a specific return matters greatly. Even if we ultimately care about the expected return, it may be beneficial during training to tolerate high variance in order to discover high reward strategies.

In this paper we introduce a risk-sensitive value function based on a system of interacting trajectories called a \emph{particle value function} (PVF).  This value function is amenable to large-scale reinforcement learning problems with nonlinear function approximation. The idea is inspired by recent advances in variational inference which bound the log marginal likelihood via importance sampling estimators \citep{iwae, vimco}, but takes an orthogonal approach to reward modifications, e.g. \citep{schmidhuber1991curious, ng1999policy}.
In Section \ref{section:risk_sensitivity}, we review risk sensitivity and a simple decision problem where risk is a consideration. In Section \ref{sec:smco}, we introduce a particle value function. In Section \ref{sec:expts}, we highlight its benefits on Cliffworld trained with policy gradients.

\section{Risk Sensitivity and Exponential Utility}
\label{section:risk_sensitivity}
We look at a finite horizon Markov Decision Process (MDP) setting where $R_t$ is the instantaneous reward generated by an agent following a non-stationary policy $\pi$, see Appendix \ref{appendix:mdp}.
A utility function $u : \R \to \R$ is an invertible non-decreasing function, which specifies a ranking over possible returns $\sum_{t=0}^T R_t$. The {expected} utility $\expect [u(\sum_{t=0}^T R_t) | S_0 = s]$ specifies a ranking over policies \citep{von1953theory}.
For an agent following $u$, a natural definition of the ``value'' of a state is the real number $V^{\pi}_T(s, u)$ whose utility is the expected utility:
\begin{align}
V^{\pi}_T(s, u) = u^{-1} \left(\expect \left[u\left(\sum_{t = 1}^T R_t\right) \ \middle | \ S_0 = s\right]\right).
\end{align}
Note, when $u$ is the identity we recover the expected return. We consider exponential utilities $u(x) = \sign(\beta) \exp(\beta x)$ where $\beta \in \R$. This choice is well-studied, and it is implied by the assumption that $V^{\pi}_T(s, u)$ is additive for deterministic translations of the reward function \citep{pratt1964risk, howard1972risk, coraluppi}. The corresponding value function is
\begin{align}
\label{eq:rsval}
V^{\pi}_T(s, \beta) = \frac{1}{\beta} \log \expect \left[\exp\left(\beta \sum_{t=0}^{T} R_t\right) \ \middle| \ S_0 = s \right],
\end{align}
and as $\beta \to 0$ we recover the expected return. See Appendix \ref{appendix:risksensitivevalue} for details. One way to interpret this value is through the following thought experiment. If the agent is given a choice between a single interaction with the environment and an immediate deterministic return, then $V^{\pi}_T(s, \beta)$ represents the minimum return that our agent would take in exchange for forgoing an interaction. If $\beta < 0$, then  $V^{\pi}_T(s, \beta) \leq V^{\pi}_T(s,0)$, meaning that the agent is willing to to take a loss relative to the expected return in exchange for certainty. This is a risk-avoiding attitude, which emphasizes low returns. If $\beta > 0$, then $V^{\pi}_T(s, \beta) \geq V^{\pi}_T(s,0)$, and the agent would only forgo an interaction for more than it can expect to receive. This is risk-seeking behavior, which emphasizes high returns.

\begin{figure}[t]
\vspace{-3\baselineskip}

    \centering
    \begin{subfigure}[b]{0.4\textwidth}
        \includegraphics[width=\textwidth]{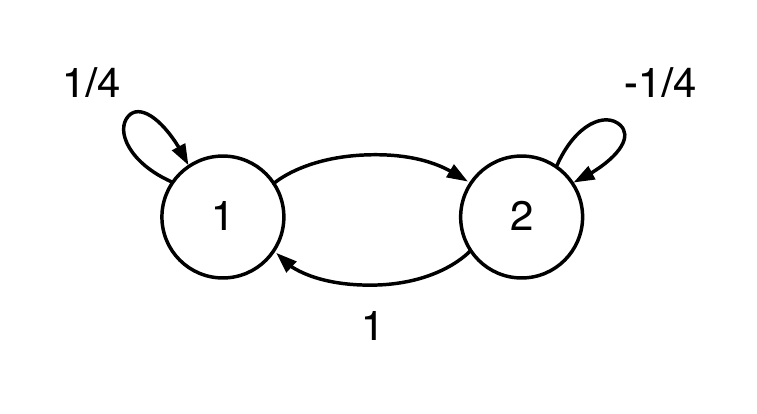}
    \end{subfigure}
    ~ %add desired spacing between images, e. g. ~, \quad, \qquad, \hfill etc. 
    \begin{subfigure}[b]{0.4\textwidth}
        \includegraphics[width=\textwidth]{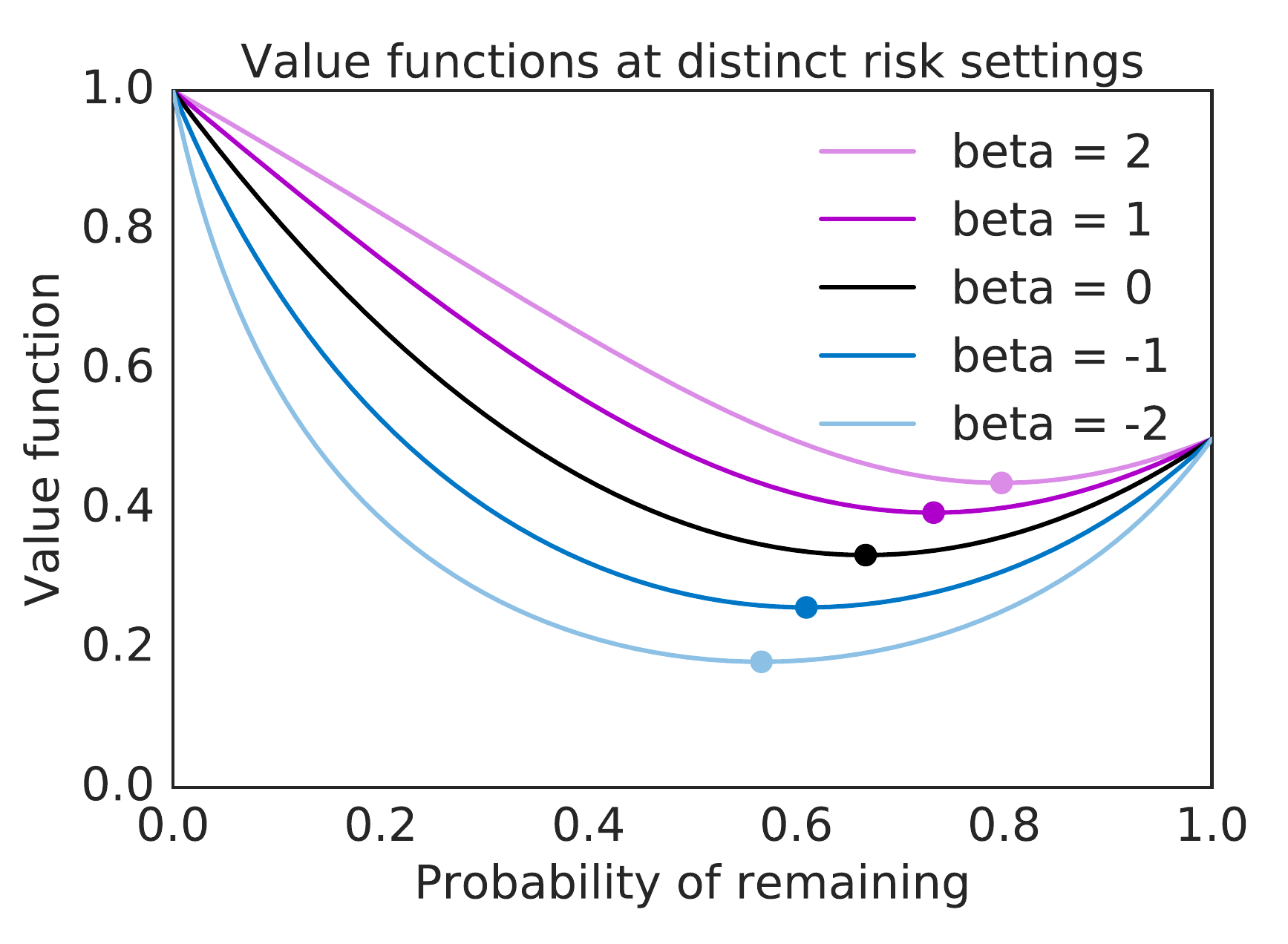}
    \end{subfigure}
    \caption{A two state MDP. The plot shows $V^{p}_2(1, \beta)$ for distinct $\beta$, assuming aliased states and a policy parameterized simply by the probability $p \in [0,1]$ of remaining in a state. }
    \label{fig:example}

\vspace{-\baselineskip}
\end{figure}

To illustrate one effect of risk, consider the two state MDP shown in Figure \ref{fig:example}. The agent begins in state 1 and acts for two time steps, choosing between leaving or remaining. Suppose that the agent's policy is defined by a single parameter $p \in [0,1]$ that describes the probability of remaining. Then the expected return $V^p_2(1, 0) = \frac{3}{2}p^2 - 2p + 1$ has two local maxima at $p \in \{0, 1\}$ and the solution $p = 1$ is not a global maximum. Any policy gradient trajectory initialized with $p > 2/3$ will converge to the suboptimal solution $p=1$, but as our risk appetite $\beta$ grows, the basin of attraction to the global maximum of $p= 0$ expands to the entire unit interval. 
This sort of state aliasing happens often in reinforcement learning with non-linear function approximation. In these cases, modifying the risk appetite (either towards risk-avoidance or seeking) may favorably modify the convergence of policy gradient algorithms, even if our ultimate objective is the expected return. 

The risk-seeking variant may be helpful in deterministic environments, where an agent can exactly reproduce a previously experienced trajectory. Rare rewards are rare only due to our current policy, and it may be better to pursue high yield trajectories more aggressively. Note, however, that $V^{\pi}_T(s, \beta)$ is non-decreasing in $\beta$, so in general risk-seeking is not guaranteed to improve the expected return. Note also that the literature on KL regularized control  \citep{todorov2006linearly, kappen2005linear, tishby2011information} gives a different perspective on risk sensitive control, which mirrors the relationship between variational inference and maximum likelihood. See Appendix \ref{appendix:relwork} for related work.

\section{Particle Value Functions}
\label{sec:smco}

Algorithms for optimizing $V^{\pi}_T(s, \beta)$ may suffer from numerical issues or high variance, see Appendix \ref{appendix:risksensitivevalue}. Instead we define a value function that bounds $V_T^{\pi}(s, \beta)$ and approaches it in the infinite sample limit. We call it a particle value function, because it assigns a value to a bootstrap particle filter with $\numagents$ particles representing state-action trajectories. This is distinct, but related to \citet{kantas2009sequential}, which investigates particle filter algorithms for infinite horizon risk-sensitive control.

Briefly, a bootstrap particle filter can be used to estimate normalizing constants in a hidden Markov model (HMM). Let $(X_t, Y_t)$ be the states of an HMM with transitions $X_t \sim p( \cdot | X_{t-1})$ and emissions $Y_t \sim q( \cdot | X_t)$. Given a sample $y_0 \ldots y_T$, the probability $p(\{Y_t = y_t\}_{t=0}^{T})$ can be computed with the forward algorithm. The bootstrap particle filter is a stochastic procedure for the forward algorithm that avoids integrating over the state space of the latent variables. It does so by propagating a set of $\numagents$ particles $X_t^{(i)}$ with the transition model $X_t^{(i)} \sim p(\cdot | X_{t-1}^{(i)})$ and a resampling step in proportion to the potentials $q(y_t | X_t^{(i)})$. The result is an unbiased estimator $\prod_{t=0}^T (\numagents^{-1} \sum_{i=1}^{\numagents} q(y_t | X_t^{(i)}))$ of the desired probability \citep{del2004feynman, pitt2012some}. The insight is that if we treat the state-action pairs $(S_t, A_t)$ as the latents of an HMM with emission potentials $\exp(\beta R_t(S_t, A_t))$ \citep[similar to][]{toussaint2006probabilistic, rawlik2010approximate}, then a bootstrap particle filter returns an unbiased estimate of $\expect[\exp(\beta \sum_{t=0}^T R_t) | S_0 = s]$. Algorithm \ref{alg:smcagent} summarizes this approach.

\begin{algorithm}[H]
\caption{An estimator of the PVF $V_{T,K}^{\pi}(s^{(1)}, \ldots, s^{(K)}, \beta)$}
\label{alg:smcagent}
  \begin{varwidth}[t]{0.4\textwidth}        % change 0.45 to suit your need
    \begin{algorithmic}[1]
    \For{$i = 1: \numagents$}
        \State $S_0^{(i)} = s^{(i)}$
        \State $A_0^{(i)} \sim \pi_T(\cdot | s^{(i)})$ 
        \State $W_0^{(i)} = \exp( \beta R_0^{(i)}) \qquad \qquad$
    \EndFor
    \State $Z_0 = \frac{1}{\numagents} \sum_{i=1}^{\numagents} W_0^{(i)}$
    
    \For{$t = 1 : T$}
        \For{$i = 1 : \numagents$}
    \algstore{myalg}
    \end{algorithmic}
  \end{varwidth}
  \begin{varwidth}[t]{0.6\textwidth}            % change 0.4 to suit your need
    \begin{algorithmic}[1]
    \algrestore{myalg}
      \State $I \sim \proba(I = j) \propto W_{t-1}^{(j)}$ \# select random parent
      \label{line:resample}
      \State $S_t^{(i)} \sim p(\cdot | S_{t-1}^{(I)}, A_{t-1}^{(I)})$ \# inherit from parent
      \State $A_t^{(i)} \sim \pi_{T-t}(\cdot | S_t^{(i)})$
      \State $W_t^{(i)} = \exp( \beta R_t^{(i)})$
    \EndFor
    \State $Z_t = \frac{1}{\numagents} \sum_{i=1}^{\numagents} W_t^{(i)}$
    \EndFor
    \State \Return $\frac{1}{\beta}\sum_{t=0}^T \log Z_t$
    \end{algorithmic}
  \end{varwidth}
\end{algorithm}

\vspace{-\baselineskip}

Taking an expectation over all of the random variables not conditioned on we define the PVF associated with the bootstrap particle filter dynamics:
\begin{align}
    V_{T}^{\pi}(s^{(1)}, \ldots, s^{(K)}, \beta) = \expect\left[\frac{1}{\beta} \sum_{t=0}^T \log Z_t \ \middle| \ \left\{S_0^{(i)} = s^{(i)}\right\}_{i=1}^{\numagents}\right].
\end{align}
Note, more sophisticated sampling schemes, see \citet{doucet2011tutorial}, result in distinct PVFs.

 Consider the value if we initialize all particles at $s$, $V_{T,K}^{\pi}(s, \beta) = V_{T}^{\pi}(s, \ldots, s, \beta)$. If $\beta > 0$, then by Jensen's inequality and the unbiasedness of the estimator we have that $V_{T,K}^{\pi}(s, \beta) \leq V_T^{\pi}(s, \beta)$. For $\beta < 0$ the bound is in the opposite direction. It is informative to consider the behaviour of the trajectories for different values of $\beta$. For $\beta > 0$ this algorithm greedily prefers trajectories that encounter large rewards, and the aggregate return is a per time step soft-max. For $\beta < 0$ this algorithm prefers trajectories that encounter large negative rewards, and the aggregate return is a per time step soft-min.  See Appendix \ref{appendix:algorithm} for the Bellman equation and policy gradient of this PVF. 

\section{Experiments}
\label{sec:expts}
\begin{figure}[b]
\vspace{-\baselineskip}
    
    \centering
        \includegraphics[width=0.75\textwidth]{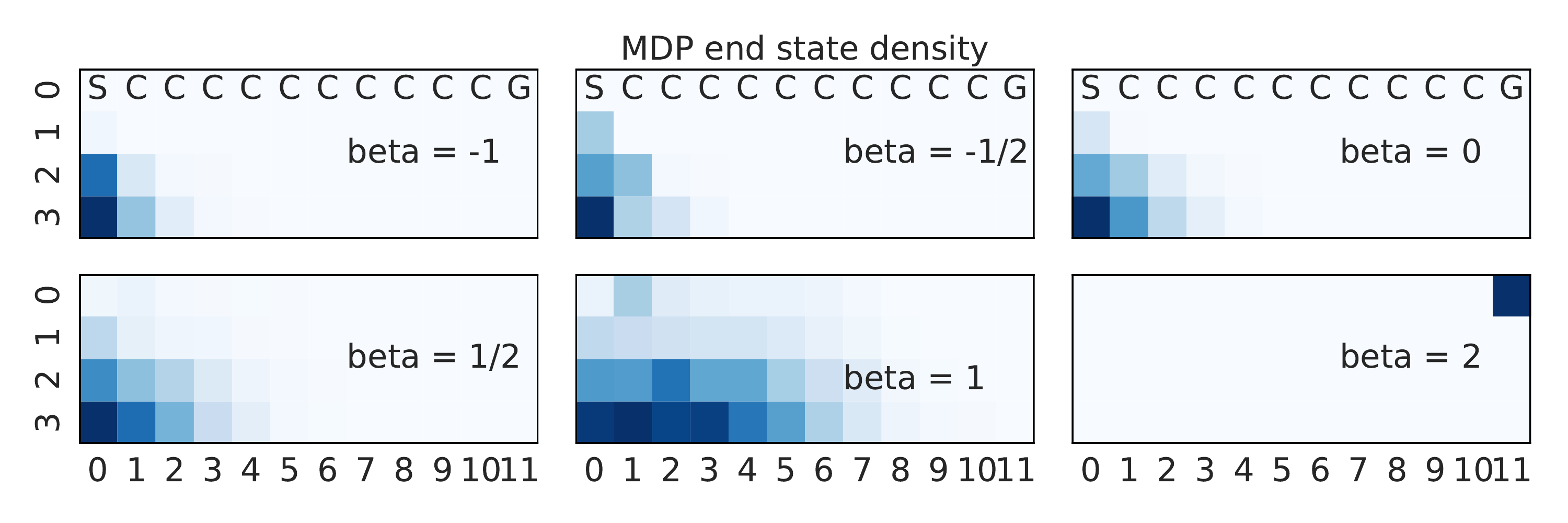}
        \caption{Last state distribution under policies trained with PVFs with distinct $\beta$.}
    \label{fig:cliffworldexpts}

\vspace{-\baselineskip}
\end{figure}

To highlight the benefits of using PVFs we apply them to a variant of the Gridworld task called Cliffworld, see Appendix \ref{appendix:cliffworld} for comparison to other methods and more details. We trained time dependent tabular policies using policy gradients from distinct PVFs for $\beta \in \{-1, -0.5, 0, 0.5, 1, 2\}$. We tried $\numagents \in \{1, \ldots, 8\}$ and learning rates $\epsilon \in \{\num{1e-3}, \num{5e-4}, \num{1e-4}, \num{5e-5}\}$. For the $\beta = 0$ case we ran $\numagents$ independent non-interacting trajectories and averaged over a policy gradient with estimated baselines. Figure \ref{fig:cliffworldexpts} shows the density over the final state of the trained MDP under varying $\beta$ treatments but $\numagents = 4$. Notice that the higher the risk parameter, the broader the policy, with the agent eventually solving the task. No $\beta = 0$, corresponding to standard REINFORCE, runs solved this task, even after increasing the number of agents to 64.

\section{Conclusion}

We introduced the particle value function, which approximates a risk-sensitive value function for a given MDP. We will seek to address theoretical questions, such as whether the PVF is increasing in $\beta$ and monotonic in the number of particles. Also, the PVF does not have an efficient tabular representation, so understanding the effect of efficient approximations would be valuable.
Experimentally, we hope to explore these ideas for complex sequential tasks with non-linear function approximators. One obvious example of such tasks is variational inference over a sequential model.

\subsubsection*{Acknowledgements}
We thank R\'{e}mi Munos, Theophane Weber, David Silver, Marc G. Bellemare, and Danilo J. Rezende for helpful discussion and support in this project.

\bibliography{iclr2017_workshop}
\bibliographystyle{iclr2017_workshop}

\newpage

\appendix

\section{Markov Decision Processes}
\label{appendix:mdp}

We consider decision problems in which an agent selects actions and receives rewards in a stochastic environment. For the sake of exposition, we consider a finite horizon MDP, which consists of: a finite state space $\mathcal{S}$, a finite action space $\mathcal{A}$, a stationary environmental transition kernel satisfying the Markov property $p( \cdot | S_t, A_t, \ldots, S_0, A_0) = p(\cdot | S_t, A_t)$, and reward functions $r_{T-t} : \mathcal{S} \times \mathcal{A} \to \R$. At each time step the agent chooses actions according to a policy $\pi_{T-t}(\cdot | S_t)$ given the current state.  $\pi_{T-t}(\cdot | S_t)$ is the action distribution and $r_{T-t}$ the reward function when there are $T-t$ steps remaining. All together the MDP proceeds stochastically producing a sequence of random variables $(S_t, A_t)$ according to the following dynamics for $T \in \mathbb{N}$ time steps. Let $t \in \{0, \ldots, T\}$,
\begin{align}
S_0 &= s \\
A_t &\sim \pi_{T-t}(\cdot | S_t) \\
S_{t+1} &\sim p(\cdot | S_t, A_t)
\end{align}
The agent receives a reward $R_t = r_{T-t}(S_t, A_t)$ at each time step. We will call a single realization of the MDP a trajectory. The objective in classical reinforcement learning is to discover the policies $\pi = \{\pi_t(\cdot | s)\}_{t=0}^T$ that maximize the value function,
\begin{align}
\label{eq:rnval}
V^{\pi}_T(s) = \expect \left[\sum_{t=0}^T R_t \ \middle| \ S_0 = s \right].
\end{align}
where the expectation is taken with respect to all the stochastic elements not conditioned on. 

All of the results we present can be simply extended to the infinite horizon case with discounted or episodic returns as well as more general uncountable state and action spaces.

\section{Risk-Sensitive Value Function Details}
\label{appendix:risksensitivevalue}

Utility theory gives us a language for describing the relative importance of high or low returns. A utility function $u : \R \to \R$ is an invertible non-decreasing function, which specifies a ranking over possible returns $\sum_{t=0}^T R_t$. The {expected} utility $\expect [u(\sum_{t=0}^T R_t) | S_0 = s]$ specifies a ranking over policies \citep{von1953theory}. The expected utility does not necessarily have an interpretable scale, because any affine transformation of the utility function results in the same relative ordering of policies or return outcomes. Therefore we define the value associated with a utility $u$ by returning it to the scale of the rewards defined by the MDP. For an agent following $u$, the ``value'' of a state is the real number $V^{\pi}_T(s, u)$ whose utility is the expected utility:
\begin{align}
V^{\pi}_T(s, u) = u^{-1} \left(\expect \left[u\left(\sum_{t = 1}^T R_t\right) \ \middle | \ S_0 = s\right]\right).
\end{align}
Note that when $u$ is the identity we recover the expected return. Of course for non-decreasing invertible utilities, the value gives the same ranking over policies. One way to interpret this value is through the following thought experiment. If the agent is given a choice between a single interaction with the environment or an immediate deterministic return, then $V^{\pi}_T(s, u)$ represents the minimum return that our agent would take in exchange for forgoing an interaction. If
$$V^{\pi}_T(s, u) \leq \expect \left[\sum_{t=0}^T R_t \middle | S_0 = s\right]$$
then our agent is willing to take a loss relative to the expected return in exchange for certainty. This is a risk-avoiding attitude, which emphasizes the smallest returns, and one can show that this occurs iff $u$ is concave. If
$$V^{\pi}_T(s, u) \geq \expect \left[\sum_{t=0}^T R_t \middle | S_0 = s\right]$$
then the agent would only forgo an interaction for more than it can expect to receive. This is risk-seeking behavior, which emphasizes the largest returns, and one can show that this occurs iff $u$ is convex. The case when $u(x)$ is linear is the risk-neutral case. For these reasons, $V^{\pi}_T(s, u)$ is also known as the \emph{certain equivalent} in economics \citep{howard1972risk}.

We focus on exponential utilities of the form $u(x) = \sign(\beta) \exp(\beta x)$ where $\beta \in \R$. This is a broadly studied choice that is implied by the assumption that the value function $V^{\pi}_T(s, u)$ is additive for deterministic translations of the return \citep{pratt1964risk, howard1972risk, coraluppi}. This assumption is nice, because it preserves the Markov nature of the decision process: if the agent were given a choice at \emph{every} time step $t$ between continuing the interaction or terminating and taking its value as a deterministic return, then additivity in the value function means that the same decision is made regardless of the return accumulated so far \citep{howard1972risk}. The value function corresponding to an exponential utility is
\begin{align}
\label{eq:rsval}
V^{\pi}_T(s, \beta) = \frac{1}{\beta} \log \expect \left[\exp\left(\beta \sum_{t=0}^{T} R_t\right) \ \middle| \ S_0 = s \right],
\end{align}
and as $\beta \to 0$ we recover the expected return. We list a few of its properties.
\begin{enumerate}
\item For $\beta$ near $0$
\begin{align}
\label{eq:rsval}
V^{\pi}_T(s, \beta) =  \expect \left[\sum_{t=0}^T R_t \ \middle| \ S_0 = s \right] + \frac{\beta}{2}  \var_{\pi} \left[\sum_{t=0}^T R_t \ \middle| \ S_0 = s \right] + o(\beta^2)
\end{align}
\item $\lim_{\beta \to \infty} V^{\pi}_T(s, \beta) = \sup \{r | \proba(\sum_{t=0}^T R_t = r) > 0 \} $
\item $\lim_{\beta \to -\infty} V^{\pi}_T(s, \beta) = \inf \{r | \proba(\sum_{t=0}^T R_t = r) > 0\} $
\item $V^{\pi}_T(s, \beta)$ is continous and non-decreasing in $\beta$.
\item $V^{\pi}_T(s, \beta)$ is risk-seeking for $\beta > 0$, risk-avoiding for $\beta = 0$, and risk-neutral for $\beta = 0$
\end{enumerate}
For proofs,
\begin{enumerate}
\item From \citet{coraluppi}.
\item From \citet{coraluppi}.
\item From \citet{coraluppi}.
\item $V^{\pi}_T(s, \beta)$ is clearly continuous for all $\beta \neq 0$. If we extend $V^{\pi}_T(s, 0) \equiv \expect \left[\sum_{t=0}^T R_t \ \middle| \ S_0 = s \right]$, then 1. gives us the continuity everywhere. For non-decreasing let $\alpha, \beta \in \R$ and $\alpha \neq 0$ and $\beta \neq 0$. Furthermore assume $\beta \geq \alpha$. Then $\beta / \alpha > 0$ or $\beta / \alpha \geq 1$. Now,
\begin{align*}
\expect\left[\exp\left(\beta\sum_{t=0}^T R_t \right) \ \middle| \ S_0 = s \right] &= \expect\left[\exp\left(\alpha \sum_{t=0}^T R_t \right)^{\beta / \alpha} \ \middle| \ S_0 = s \right]\\
&\geq \expect\left[\exp\left(\alpha \sum_{t=0}^T R_t \right) \ \middle| \ S_0 = s \right]^{\beta / \alpha}
\end{align*}
since $x^p$ is convex on $x > 0$ for $p \geq 1$ or $p < 0$, Jensen's inequality gives us the result. Taking log of both sides gives us the result in that case. In the case that $\alpha = 0$ or $\beta = 0$, 4. and Jensen's inequality gives us the result by the concavity of log.
\end{enumerate}

From a practical point of view the value function $V^{\pi}_T(s, \beta)$ behaves like a soft-max or soft-min depending on the sign of $\beta$, emphasizing the avoidance of low returns when $\beta < 0$ and the pursuit of high returns when $\beta > 0$. As $\beta \to \infty$ the value $V^{\pi}_T(s, \beta)$ approaches the supremum of the returns over all trajectories with positive probability, a best-case penalty. As $\beta \to -\infty$ it approaches the infimum, a worst-case value \citep{coraluppi}. Thus for large positive $\beta$ this value is tolerant of high variance if it can lead to high returns. For large negative $\beta$ it is very intolerant of rare low returns.

Despite having attractive properties the risk-sensitive value function is not always applicable to reinforcement learning tasks (see also \citet{mihatsch2002risk}). The value function satisfies the multiplicative Bellman equation
\begin{align}
\exp(\beta V^{\pi}_T(s, \beta)) = \sum_{a, s^{\prime}} \pi_T(a | s) p(s^{\prime} | s, a) \exp(\beta r_T(s, a) + \beta V^{\pi}_{T-1}(s, \beta)).
\end{align}
Operating in $\log$-space breaks the ability to exploit this recurrence from Monte Carlo returns generated by a single trajectory, because expectations do not exchange with $\log$. Operating in $\exp$-space is possible for TD learning algorithms, but we must cap the minimum/maximum possible return so that $\exp(\beta V^{\pi}_T(s, \beta))$ does not underflow/overflow. This can be an issue when the rewards represent log probabilities as is often the case in variational inference. The policy gradient of $V^{\pi}_T(s, \beta)$ is
\begin{align}
\nabla_{\pi}V^{\pi}_T(s, \beta) = \expect\left[\frac{1}{\beta}\sum_{t=0}^T \exp(\beta Q^{\pi}_{T-t}(S_t, A_t, \beta)  - \beta V^{\pi}_T(S_t, \beta) ) \nabla \log \pi_{T-t}(A_t | S_t) \middle | S_0 = s\right]
\end{align}
where
\begin{align}
Q^{\pi}_{T}(s, a, \beta) = \frac{1}{\beta} \log \expect \left[ \exp \left(\beta \sum_{t=0}^T R_t \right)\middle | S_0 = s, A_0 = a\right]
\end{align}
Even ignoring underflow/overflow issues, REINFORCE \citep{williams1992simple} style algorithms would find difficulties, because deriving unbiased estimators of the ratio $\exp(\beta Q^{\pi}_{T-t}(S_t, A_t, \beta)  - \beta V^{\pi}_T(s, \beta))$ from single trajectories of the MDP may be hard. Lastly, the policy gradient of $\expect \left[\exp\left(\beta \sum_{t=0}^{T} R_t\right) \ \middle| \ S_0 = s \right]$
\begin{align}
\nabla_{\pi} \expect \left[\exp\left(\beta \sum_{t=0}^{T} R_t\right) \ \middle| \ S_0 = s \right] = \expect\left[\sum_{t=0}^T\exp\left(\beta \sum_{t=0}^T R_t\right)\nabla \log \pi_{T-t}(A_t | S_t) \middle | S_0 = s\right],
\end{align}
There are particle methods that would address the estimation of this score, e.g. \citep{kantas2015particle}, but for large $T$ the estimate suffers from high mean squared errors.

\section{Related Work}
\label{appendix:relwork}

Risk sensitivity originates in the study of utility and choice in economics \citep{von1953theory, pratt1964risk, arrow1974essays}. It has been extensively studied for the control of MDPs \citep{howard1972risk, coraluppi, marcus1997risk, borkar2002risk, mihatsch2002risk, bauerle2013more}. In reinforcement learning, risk sensitivity has been studied \citep{koenig1994risk, neuneier1998risk, shen2014risk}, although none of these consider the direct policy gradient approach considered in this work. Most of the methods considered are variants of a Q learning approach or policy iteration. As well, the idea of treating rewards as emissions of an HMM is not a new idea \citep{toussaint2006probabilistic, rawlik2010approximate}.

The idea of treating reinforcement learning as an inference problem is not a new idea \citep[see e.g.][]{albertini1988logarithmic, dayan1997using, kappen2005linear, toussaint2006probabilistic, hoffman2007solving, tishby2011information, kappen2012optimal}. Broadly speaking, all of these works still optimize the expected reward objective $V^{\pi}_T(s, 0) = \expect[\sum_{t=0}^T R_t \ | \ S_0 = s]$ with or without some regularization penalties on the policy. The ones that share the closest connection to the risk sensitive objective $V^{\pi}_T(s, \beta)$ studied here, are the KL regularized objectives of the form
\begin{align}
\hat{V}^{\pi, \pi^{\prime}}_T(s, \beta) = \expect_{\pi^{\prime}} \left[ \sum_{t=0}^T R_t + \frac{1}{\beta} \log \frac{\pi(A_t | S_t) }{\pi^{\prime}(A_t | S_t)} \ \middle | \ S_0 = s\right] 
\end{align}
where the MDP dynamics are sampled from $\pi^{\prime}$. These are studied for example in \citet{albertini1988logarithmic, kappen2005linear, tishby2011information, kappen2012optimal, fox2015taming, ruiz2016particle}. The observation is that in an MDP with \emph{fully controllable} transition dynamics, optimizing a policy $\pi^{\prime}$, which completely specifies the transition dynamics, achieves the risk sensitive value at $\pi$:
\begin{align}
\label{eq:rlvi}
\max_{\pi^{\prime}} \hat{V}^{\pi, \pi^{\prime}}_T(s, \beta) = V^{\pi}_T(s, \beta)
\end{align}
Note that this has an interesting connection to Bayesian inference. Here, $\pi$ plays the role of the prior, $\pi^{\prime}$ the role of the variational posterior, $\hat{V}^{\pi, \pi^{\prime}}_T(s, \beta) $ the role of the variational lower bound, and $V^{\pi}_T(s, \beta)$ the role of the marginal likelihood. In effect, KL regularized control is like variational inference, where risk sensitive control is like maximum likelihood. Finally, when the environmental dynamics $p(\cdot | s, a)$ are stochastic, (\ref{eq:rlvi}) does not \emph{necessarily} hold, therefore the risk sensitive value is distinct in this case. Yet, in certain special cases, risk sensitive objectives can also be cast as solutions to path integral control problems \citep{broek2012risk}.

To our knowledge no work has considered using particle filters for risk sensitive control by treating the particle filter's estimator of the log partition function as a return whose expectation bounds the risk sensitive value and whose policy gradients are cheap to compute.

\section{Particle Value Function Details}
\label{appendix:algorithm}

Recalling Algorithm \ref{alg:smcagent} and the definition of the MDP in Appendix \ref{appendix:mdp}, define
\begin{align}
R_t^{(i)} = r_{T-t}(S_t^{(i)}, A_t^{(i)})
\end{align}
the particle value function associated with the bootstrap particle filter dynamics:
\begin{align}
V^{\pi}_T(s^{(1)}, \ldots, s^{(\numagents)}, \beta) = \expect \left[\frac{1}{\beta} \sum_{t=0}^{T} \log Z_t \ \middle| \ \left\{ S_0^{(i)} = s^{(i)} \right\}_{i=1}^K \right].
\end{align}
We can also think of this value function as the expected return of an agent whose actions space is the product space $\mathcal{A}^{\numagents}$, in an environment with state space $\mathcal{S}^{\numagents}$ whose transition kernel includes the resampling dynamic. Let $s^{(1:\numagents)} = (s^{(1)}, \ldots, s^{(\numagents)})$, then the PVF satisfies the Bellman equation
\begin{align}
\beta V^{\pi}_T(s^{(1:\numagents)}, \beta) = &\sum_{a^{(1:\numagents)}} \Pi_T(a^{(1:\numagents)} | s^{(1:\numagents)}) \log Z_{T}(a^{(1:\numagents)}, s^{(1:\numagents)}) \ + \\
& \sum_{a^{(1:\numagents)}} \sum_{\sigma^{(1:\numagents)}} \Pi_T(a^{(1:\numagents)} | s^{(1:\numagents)}) P_T(\sigma^{(1:\numagents)} | a^{(1:\numagents)}, s^{(1:\numagents)}) \beta V^{\pi}_{T-1}(\sigma^{(1:\numagents)}, \beta)
\end{align}
where
\begin{align}
\Pi_T(a^{(1:\numagents)} | s^{(1:\numagents)}) &= \prod_{i=1}^K \pi_{T}(a^{(i)} | s^{(i)})\\
\log Z_T(a^{(1:\numagents)}, s^{(1:\numagents)}) &= \log \left(\frac{1}{K}\sum_{i=1}^K \exp(\beta r_T(s^{(i)}, a^{(i)}))\right)\\
P_T(\sigma^{(1:K)} | a^{(1:\numagents)} , s^{(1:\numagents)}) &= \prod_{i=1}^K \left( \sum_{j=1}^K \frac{\exp(\beta r_T(s^{(j)}, a^{(j)}))}{\sum_{k=1}^K \exp(\beta r_T(s^{(k)}, a^{(k)}))} p(\sigma^{(i)} | s^{(j)}, a^{(j)}) \right)
\end{align}
The policy gradient of $V^{\pi}_T(s^{(1:\numagents)}, \beta)$ is
\begin{align}
\nabla_{\pi} V^{\pi}_T(s^{(1:\numagents)}, \beta) = \expect \left[\frac{1}{\beta} \sum_{t=0}^{T} \sum_{t^{\prime} = t}^T \sum_{i=1}^{\numagents} \log Z_{t^{\prime}} \nabla \log \pi_{T-t}(A_t^{(i)} | S_t^{(i)}) \ \middle| \ \{S_0^{(i)} = s^{(i)}\}_{i=1}^K \right]
\end{align}
In this sense we can think of $\log Z_t / \beta$ as the immediate reward for the whole system of particles and $\sum_{t=0}^T \log Z_t / \beta$ as the return.

The key point is that the use of interacting trajectories to generate the Monte Carlo return ensures that this particle value function defines a bound on $V_T^{\pi}(s, \beta)$. Indeed, consider the particle value function that corresponds to initializing all $\numagents$ trajectories in state $s \in \mathcal{S}$ and define, $V^{\pi}_{T,\numagents}(s, \beta) = V^{\pi}_T(s, \ldots, s, \beta)$. Now, for $\beta > 0$ we have, by Jensen's inequality,
\begin{align}
V^{\pi}_{T,\numagents}(s, \beta) &\leq \frac{1}{\beta} \log \expect \left[\prod_{t=0}^T Z_t \ \middle| \ \{S_0^{(i)} = s\}_{i=1}^K \right] \\
\intertext{and since the bootstrap particle filter is unbiased \citep{del2004feynman, pitt2012some},}
&= V^{\pi}_T(s, \beta)
\end{align}
For $\beta < 0$ we get the reverse inequality, $V^{\pi}_{T,\numagents}(s, \beta)  \geq V^{\pi}_T(s, \beta)$. As $\numagents \to \infty$ the particle value function converges to $V^{\pi}_T(s, \beta)$ since the estimator is consistent \citep{del2004feynman}. We list this and some other properties:
\begin{enumerate}
\item $V^{\pi}_{T,\numagents}(s, \beta) \leq V^{\pi}_T(s, \beta)$.
\item $\lim_{K \to \infty} V^{\pi}_{T,\numagents}(s, \beta) = V^{\pi}_T(s, \beta)$.
\item $\lim_{\alpha \to 0} V^{\pi}_{T,\numagents}(s, \alpha) = \expect \left[\sum_{t=0}^T R_t \ \middle| \ S_0 = s \right] =  V^{\pi}_{T,1}(s, \beta)$.
\item $V^{\pi}_{T,\numagents}(s, \beta)$ is continuous in $\beta$.
\end{enumerate}
For proofs,
\begin{enumerate}
\item $\prod_{t=0}^T Z_t$ is an unbiased estimator of $\expect[\exp(\beta \sum_{t=0}^T R_t) | S_0 = s]$ \citep{del2004feynman} and the rest follows from Jensen's inequality.
\item $\prod_{t=0}^T Z_t$ is a consistent estimator \citep{del2004feynman}, and the rest follows from exchanging the limit with an expectation.
\item $V^{\pi}_{T,1}(s, \beta) = \expect \left[\sum_{t=0}^T R_t \ \middle| \ S_0 = s \right]$ is clear. Otherwise the limit $\lim_{\beta \to 0} V^{\pi}_{T,K}(s, \beta)$ approaches the algorithm that resamples uniformly and the value under that sampling strategy is
\begin{align*}
\lim_{\beta \to 0}  V^{\pi}_{T,K}(s, \beta) &= \expect \left[\sum_{t=0}^T \lim_{\beta \to 0} \frac{\log Z_t}{\beta} \ \middle| \ \{ S_0^{(i)} = s\}_{i=1}^K\right]\\
&= \expect  \left[ \sum_{t=0}^T \sum_{i=1}^K \frac{1}{K} R_t^{(i)}  \ \middle| \ \{ S_0^{(i)} = s\}_{i=1}^K\right] \\
&= \sum_{t=0}^T \sum_{i=1}^K \frac{1}{K} \expect  \left[ R_t^{(i)}  \ \middle| \ \{ S_0^{(i)} = s\}_{i=1}^K\right] \\
\intertext{Because each $R_t^{(i)}$ has a genealogy, which is an MDP trajectory}
&= \sum_{t=0}^T \sum_{i=1}^K \frac{1}{K} \expect  \left[ R_t \ \middle| \ S_0 = s\right]\\
&= \expect  \left[  \sum_{t=0}^T R_t \ \middle| \ S_0 = s\right]
\end{align*}
\item $V^{\pi}_{T,\numagents}(s, \beta)$ is a finite sum of continuous terms, and if we extend the definition of $V^{\pi}_{T,\numagents}(s, 0) \equiv \lim_{\beta \to 0} V^{\pi}_{T,\numagents}(s, \beta)$, then we're done.
\end{enumerate}

\section{Cliffworld Details}
\label{appendix:cliffworld}

 \begin{figure}[t]

    \vspace{-\baselineskip}

    \centering
    \includegraphics[width=0.5\textwidth]{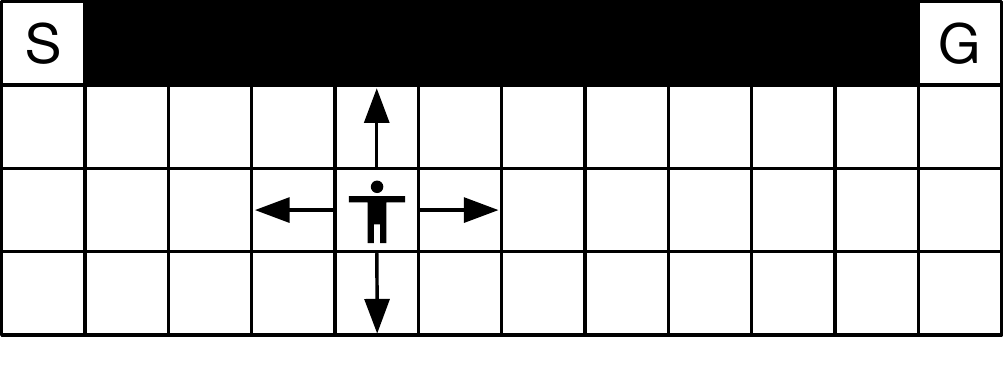}
    \caption{Cliffworld is a $n \times m$ gridworld. S denotes the start state, G the goal state, and the agent is currently in state (4,2). The arrows show the actions available to the agent.}
    \label{fig:cliffworld}
    
\end{figure}

 \begin{figure}[t]
        
    \centering
    \includegraphics[width=0.4\textwidth]{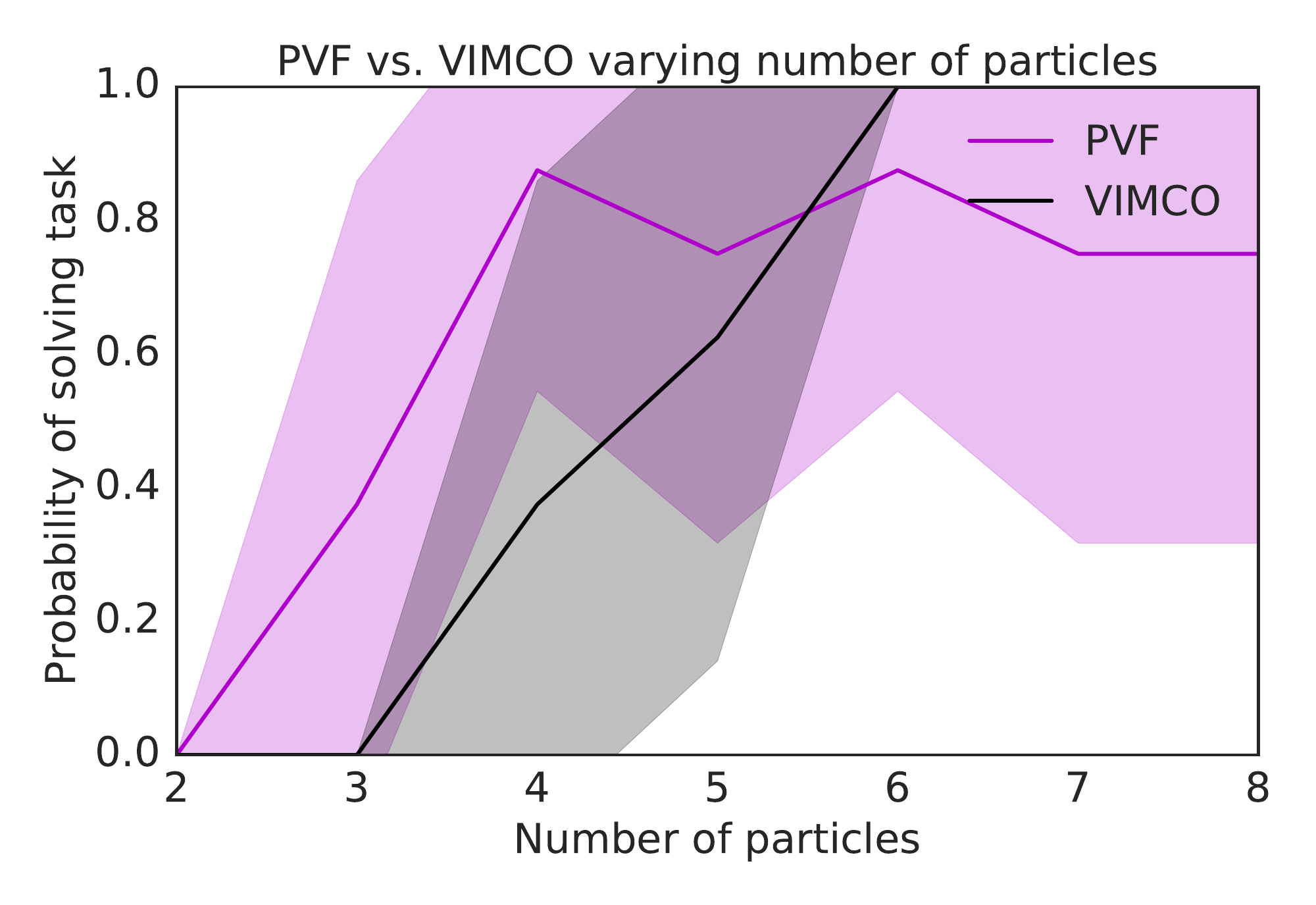}
    \includegraphics[width=0.4\textwidth]{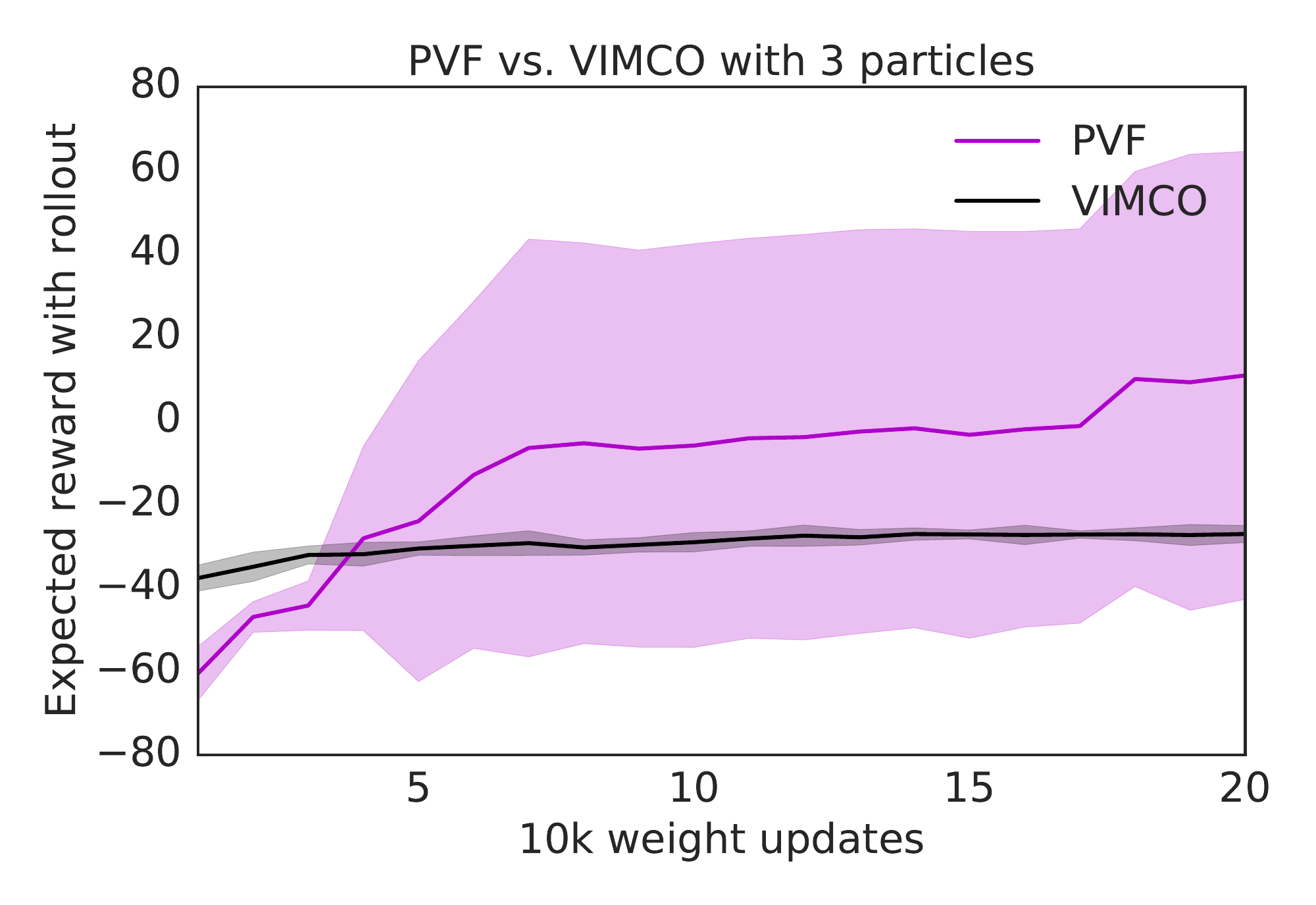}
    \caption{Left plot: probability of solving task with standard deviation, defined as achieving positive average return. Right plot: Average reward during training with standard deviation. Both VIMCO and PVF trained with $\beta = 1.0$ and learning rate $\epsilon = \num{1e-3}$. Averages are for 8 runs. At 3 particles, some PVF runs began solving Cliffworld, while no VIMCO ones did.}
    \label{fig:cliffworldres}

    \vspace{-\baselineskip}

\end{figure}

We considered a finite horizon Cliffworld task, which is a variant on Gridworld. The world is 4 rows by 12 columns, and the agent can occupy any grid location. Each episode begins with the agent in state (0,0) (marked as S in Figure \ref{fig:cliffworld}) and ends when 24 timesteps have passed. The actions available to the agent are moving north, east, south, and west, but moving off the grid is prohibited. All environmental transitions are deterministic.
 The `cliff' occupies all states between the start and the goal (marked G in Figure \ref{fig:cliffworld}) along the northern edge of the world. All cliff states are absorbing and when the agent enters any cliff state it initially receives a reward of -100 and receives 0 reward each timestep thereafter. The goal state is also absorbing, and the agent receives a +100 reward upon entering it and 0 reward after. The agent receives a -1 reward for every  action that does not transition into a cliff or goal state.
The optimal policy for Cliffworld is to hug the cliff and proceed from the start to the goal as speedily as possible, but doing so could incur high variance in reward if the agent falls off the cliff. For a uniform random policy most trajectories result in large negative rewards and occasionally a high positive reward. This means that initially for independent trajectories venturing east is high variance and low reward.
 
 We trained non-stationary tabular policies parameterized by parameters $\theta$ of size $4 \times 12 \times 4 \times 24$:
 $$\pi_{T-t}(a | s) = \frac{\exp(\theta[s_1, s_2, a, T-t])}{\sum_{a = 0}^3 \exp(\theta[s_1, s_2, a, T-t])}$$
 The policies were trained using policy gradients from distinct PVFs for $\beta \in \{-1, -0.5, 0, 0.5, 1, 2\}$. We tried $\numagents \in \{1, \ldots, 8\}$ and learning rates $\epsilon \in \{\num{1e-3}, \num{5e-4}, \num{1e-4}, \num{5e-5}\}$. For the $\beta = 0$ case we ran $\numagents$ independent non-interacting trajectories and averaged over a policy gradient with estimated baselines.  For $\beta = 0$, we used instead a  REINFORCE \citep{williams1992simple} estimator, that was simply estimated from the Monte Carlo returns. 
For control variates, we used distinct baselines depending on whether $\beta = 0$ or not. For $\beta = 0$, we used a baseline that was an exponential moving average with smoothing factor 0.8. The baselines were also non-stationary, and with dimensionality $4 \times 12 \times 24$. For $\beta \neq 0$ we used no baseline except for VIMCO's control variate \citep{vimco} for the immediate reward. The VIMCO control variate is not applicable for the whole return as future time steps are correlated with the action through the interaction of trajectories. 
 
 We also compared directly to VIMCO \citep{vimco}. Consider VIMCO's value function,
 \begin{align}
\tilde{V}^{\pi}_{T, \numagents}(s, \beta) = \expect \left[ \frac{1}{\beta}\log\left(\frac{1}{\numagents}\sum_{i=1}^{\numagents} \exp\left(\sum_{t=0}^T \beta R_t^{(i)}\right) \right)\right]
 \end{align}
where $R_t^{(i)}$ is a reward sequence generated by an independent Monte Carlo rollout of the original MDP. VIMCO is also a risk sensitive value function, but it does not decompose over time and so does not have a temporal Bellman equation. In this case, though, VIMCO policy gradients were able to solve Cliffworld under most of the conditions that the policy gradients of PVF were able to solve. For $K = 3$ and $\beta = 1.0$, PVF occasionally solved Cliffworld while VIMCO did not. See Figure \ref{fig:cliffworldres}. However, once in the regime where VIMCO could solve the task, it did so with more reliability than the PVF variant. Note that in no case did REINFORCE on the expected return solve this variant.

 \end{document}